%% file: root.tex
\documentclass[letterpaper, 10 pt, conference]{ieeeconf}  

\IEEEoverridecommandlockouts                              

\usepackage{amssymb}
\usepackage{graphicx}
\usepackage{subfigure}
\usepackage{cite}
\usepackage{graphics}
\usepackage{CJKutf8}
\usepackage{color}
\usepackage[T1]{fontenc}
\usepackage{amsmath}
\usepackage{hyperref}
\usepackage[ruled,linesnumbered]{algorithm2e}
\usepackage{algorithmic}
\hypersetup{hidelinks,
	pdfstartview=Fit,
	breaklinks=true}
\usepackage{multirow}
\usepackage{booktabs}
\usepackage{subfigure}
\usepackage{pifont}

\title{\LARGE \bf
MSGField: A Unified Scene Representation Integrating Motion, Semantics, and Geometry for Robotic  Manipulation
}

\author{Yu Sheng$^{1}$, Runfeng Lin$^{1}$,  Lidian Wang$^{1}$, Quecheng Qiu$^{1}$, YanYong Zhang$^{1,2}$\\ Yu Zhang$^{1,2}$, Bei Hua$^{1}$, and Jianmin Ji$^{1,2,\dag}$
\thanks{$^1$ School of Computer Science and Technology, University of Science and Technology of China (USTC), Hefei 230026, China}
\thanks{$^2$ Institute of Artificial Intelligence, Hefei Comprehensive National Science Center, Hefei, Anhui, China}
\thanks{$^\dag$ Corresponding author. {\tt\small jianmin@ustc.edu.cn}}
}

\begin{document}

\maketitle
\thispagestyle{empty}
\pagestyle{empty}
\newcommand{\snote}[1] {{$\langle${\textcolor{red}{sheng: \textbf{#1}}}$\rangle$}}
\newcommand{\lnote}[1] {{$\langle${\textcolor{green}{lidian: \textbf{#1}}}$\rangle$}}
\newcommand{\qnote}[1] {{$\langle${\textcolor{blue}{qqc: \textbf{#1}}}$\rangle$}}
\begin{abstract}
Combining accurate geometry with rich semantics has been proven to be highly effective for language-guided robotic manipulation.
Existing methods for dynamic scenes either fail to update in real-time or rely on additional depth sensors for simple scene editing, limiting their applicability in real-world.
In this paper, we introduce MSGField, a representation that uses a collection of 2D Gaussians for high-quality reconstruction, further enhanced with attributes to encode semantic and motion information.
Specially, we represent the motion field compactly by decomposing each primitive's motion into a combination of a limited set of motion bases.
Leveraging the differentiable real-time rendering of Gaussian splatting, we can quickly optimize object motion, even for complex non-rigid motions, with image supervision from only two camera views.
Additionally, we designed a pipeline that utilizes object priors to efficiently obtain well-defined semantics.
In our challenging dataset, which includes flexible and extremely small objects, our method achieve a success rate of 79.2$\%$ in static and 63.3$\%$ in dynamic environments for language-guided manipulation.
For specified object grasping, we achieve a success rate of 90$\%$, on par with point cloud-based methods.
Code and dataset will be released at: \href{ https://shengyu724.github.io/MSGField.github.io}{MSGField.github.io}.

\end{abstract}

\input{intro}
\input{related_work}
\input{methods}
\input{experiments}
\input{conclusions}






\bibliographystyle{IEEEtran}
\bibliography{IEEEabrv,references}

\end{document}

%% file: intro.tex
\section{INTRODUCTION}
Embodied AI has accelerated the integration of robotic systems into daily life~\cite{ACT,fang2020graspnet}. 
It's critical for robots that can interact with humans and autonomously perform household tasks in real-world unstructured environments~\cite{liu2024aligning}.
 To achieve this, the robot needs to comprehend the scene's semantics for interaction and be aware of the objects' geometry and motion within the environment for effective manipulation.

Recently, some methods~\cite{jauhriLearningAnyView6DoF2024,GeFF,wangSparseDFFSparseViewFeature2024,wangFieldsDynamic3D2023} employ point clouds to capture the geometry of dynamic scenes and project them onto images to obtain semantic features.
However, these methods often struggle with accurately representing small objects and precise semantics. 
Others~\cite{lerf-togo,shenDistilledFeatureFields2023,zhengGaussianGrasper3DLanguage2024,qin2024langsplat,shorinwa2024splat} use Neural Radiance Field (NeRF)~\cite{mildenhallNeRFRepresentingScenes2020} or 3D Gaussians Splatting (3DGS)~\cite{kerbl3DGaussianSplatting2023} to extract more detailed semantic features, leading to enhanced scene understanding.
But the geometric precision is limited, and per-scene optimization or simple scene edits make them infeasible for dynamic settings.

\begin{figure}[t]
\vspace{0.3cm}
\centering
 \begin{minipage}{1\linewidth}
    \centering
        \includegraphics[width=1\linewidth]{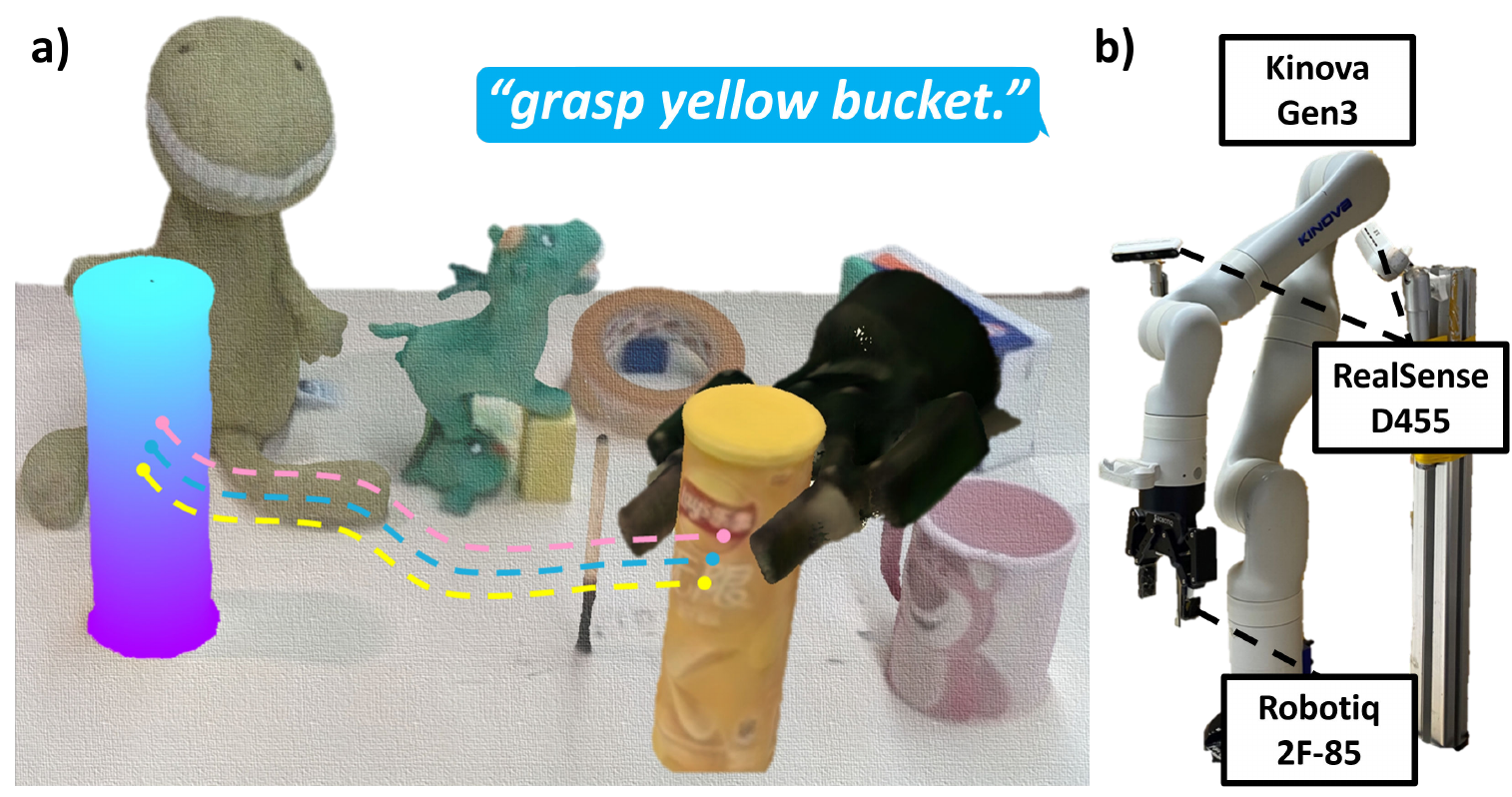}
    \end{minipage}
\caption{a).  We propose MSGField, a unified representation that enables robots to manipulate objects in real-world environments based on human instructions. b) The device we use for robotic manipulation in the real world.}
\label{msg_field}
\vspace{-2.0em}
\end{figure}

In this context, we explore the question, \textbf{"How to design a scene representation that simultaneously integrates precise motion, semantic, and geometric while remaining compact enough for rapid training?"}
In this paper, we introduce MSGField, which employs extended 2D Gaussians Splatting (2DGS)~\cite{huang2DGaussianSplatting2024} for high-fidelity surface reconstruction and provides a compact, comprehensive representation of scene motion and semantic, as shown in Fig.~\ref{msg_field}.
Unlike existing methods that directly optimize features~\cite{qin2024langsplat,zhou2024feature3dgs,kerr2023lerf} or motions~\cite{fischerDynamic3DGaussian2024a,huangSCGSSparseControlledGaussian2024,luitenDynamic3DGaussians2023,bansal20204d} on a per-point basis, we argue that such information is consistently present within local regions, rendering per-point learning unnecessary.
Specifically, for semantic filed, we design an object-centric semantic field distillation pipeline, ensuring primitives of the same object share a common semantic feature. 
This results in a lightweight yet well-defined semantic field, enabling fast training (30s) and precise 3D open-vocabulary segmentation.
For the motion, we harness a small number of learnable motion bases to capture scene dynamics, with each Gaussian primitive's trajectory represented as a linear composite of these bases.
The motion representation, tightly integrated with Gaussian primitives, allows us to leverage the differentiable real-time rendering of 2DGS for rapid optimization (5s for rigid bodies motion) with only RGB supervision, even capturing non-rigid motions that previous methods unable to describe~\cite{zhengGaussianGrasper3DLanguage2024,shorinwa2024splat}.
We collect a real-world dataset comprising 30 diverse object categories, including complex surfaces objects and small cable tie only 5mm wide. 
Our method achieved 79.2$\%$ success in static and 63.3$\%$ in dynamic environments for language-guided manipulation, and 90.0$\%$ for target-specific tasks, comparable to point cloud-based methods.

Our main contributions are as follows:
\begin{itemize}
    \item We introduce MSGField, a strategically designed scene representation that elegantly integrates motion, semantics, and geometry, while meeting the robot's need for timely response with a compact expression, significantly enhancing language-guided manipulation performance in real-world complex scenarios.
    
    \item To the best of our knowledge, MSGField is the first to apply 2DGS to robotic manipulation,  with its precise reconstruction bridging the gap between 2D and 3D sensors
    
    \item We propose a well-defined semantic field and a compact motion field that facilitate precise 3D open-vocabulary segmentation and accurate recovery complex motion of dynamic objects.
\end{itemize}

%% file: related_work.tex
\section{RELATED WORK}
\subsection{Scene Representation for Robot Manipulation}
Based on the implicit representation, F3RM~\cite{shenDistilledFeatureFields2023}, NeuGraspNet~\cite{jauhriLearningAnyView6DoF2024} and LERF-TOGO~\cite{lerf-togo} utilize NeRF to reconstruct target scenes and employ feature distillation to achieve language-guided grasping. GeFF~\cite{GeFF} utilizes pixelNeRF's~\cite{yu2021pixelnerf} single-frame reconstruction capability to achieve fast mapping.
For explicit representation,~\cite{wangSparseDFFSparseViewFeature2024,wangFieldsDynamic3D2023} accomplish one-shot manipulation by back-projecting panoramic RGB-D images and their corresponding image features into 3D space. 
GaussianGrasper~\cite{zhengGaussianGrasper3DLanguage2024} models the environment using 3DGS, where each Gaussian primitive is associated with a feature attribute to learn corresponding semantic features. 
However, per-scene optimization limits the applicability of these methods in dynamic scenes.
Although GeFF can reconstruct scenes swiftly, its mapping paradigm struggles with rapid scene changes, and the precision of single-frame reconstruction is limited.
We leverage  geometrically accurate radiance fields, 2DGS~\cite{huang2DGaussianSplatting2024}, combined with a compact motion field representation to address the aforementioned issues.

\subsection{Semantic Field Distilling}
Most of existing semantic field distilling methods~\cite{qin2024langsplat,kerr2023lerf,lerf-togo,shenDistilledFeatureFields2023,zhengGaussianGrasper3DLanguage2024,zhou2024feature3dgs} use CLIP~\cite{CLIP} to generate semantic feature maps and supervise NeRF or 3DGS models with per-pixel feature loss. 
Others~\cite{wangSparseDFFSparseViewFeature2024,wangFieldsDynamic3D2023}  use RGB-D cameras to back-project feature maps into 3D space.
However, CLIP extracts object or patch level features shared across regions, while current methods redundantly learn features point by point, leading to high memory consumption, especially in explicit representation-based approaches~\cite{qin2024langsplat,zhengGaussianGrasper3DLanguage2024,zhou2024feature3dgs}.
\cite{tziafas20243dfd} uses object-centric priors to maintain semantic consistency but relies on extensive pre-training, making it hard to combine with exist scene representation. 
In contrast, our approach achieves similar results with significantly lower time and space complexity.

\subsection{Dynamic Reconstruction and Tracking}
Dynamic reconstruction recovers 3D scenes at each timestamp, enabling the tracking of both rigid and non-rigid motions, whereas traditional tracking or pose estimation methods~\cite{huang2023earl,wen2024foundationpose} are limited to rigid bodies.
Previous dynamic reconstruction techniques relied on panoramic RGB-D cameras~\cite{bozic2020deepdeform,dou2016fusion4d} or manual priors~\cite{kumar2017monocular}. 
Recently, methods based on NeRF and 3DGS have achieved state-of-the-art results.
However, most of these approaches~\cite{fischerDynamic3DGaussian2024a,huangSCGSSparseControlledGaussian2024,luitenDynamic3DGaussians2023,bansal20204d,cao2023hexplane,broxton2020immersive} require panoramic cameras or RGB-D cameras around scene, which are challenging to implement on robots.
Some works~\cite{caiGSPoseCascadedFramework2024,wangShapeMotion4D2024} have demonstrated remarkable dynamic reconstruction using monocular images or RGB-D sequences. 
Yet, the several hours optimization hinders their real-world utility. 
In contrast, our method
achieves motion tracking in seconds using only two camera views, offering a practical solution for robotics.

%% file: methods.tex
\section{METHODS}
\label{methods}
\begin{figure*}[t]
\vspace{0.3cm}
\centering
 \begin{minipage}{1\linewidth}
    \centering
        \includegraphics[width=1\linewidth]{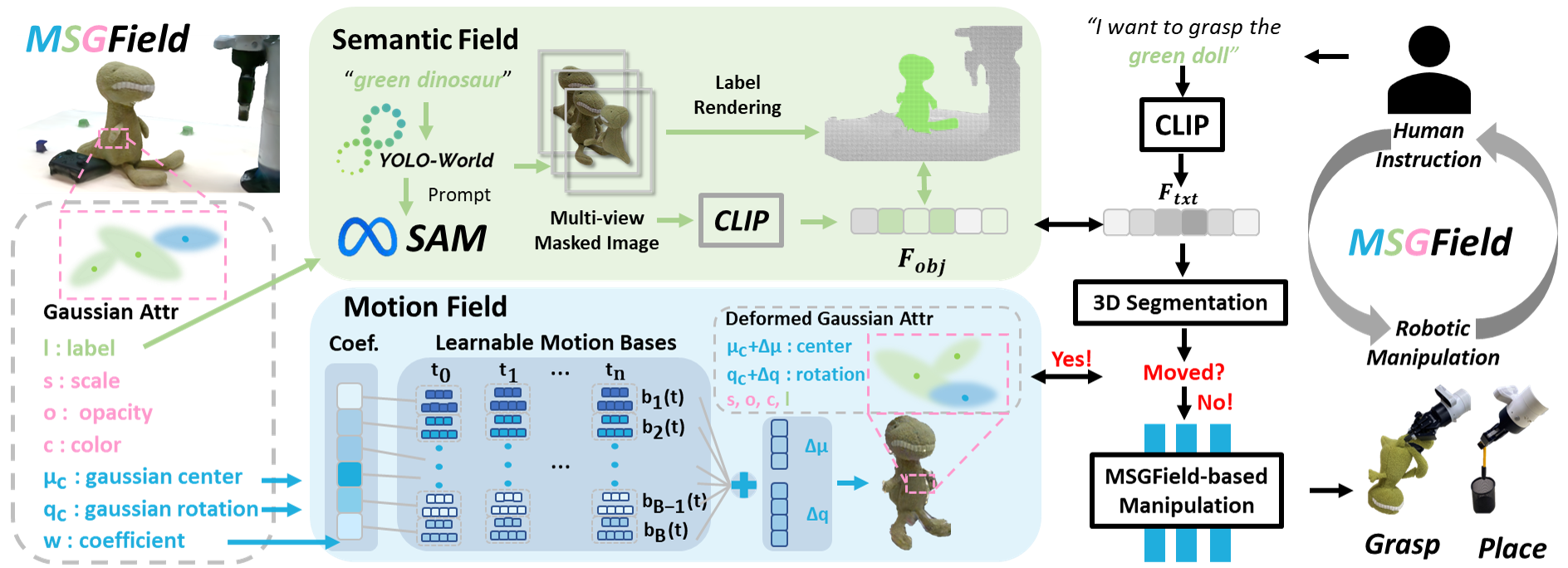}
    \end{minipage}
\caption{The framework of MSGField. Geometry field captured by surface reconstruction from 2D Gaussian Splatting. In the semantic field, each primitive is assigned a label, which links to an object feature extract from CLIP. For the motion field, we represent scene motion with motion bases, where each primitive's motion is a combination of these bases. With human instructions, objects are segmented via text queries, and their motion is tracked. A grasp detector then identifies executable manipulations.}
\label{pipeline}
\vspace{-2em}
\end{figure*}

In this sections, we first introduce preliminaries on Gaussian Splatting and detail why 2D Gaussians ensure precise geometric reconstruction. 
Then we discuss how the MSGField efficiently and precisely represents scene semantics and motions.
Finally, we demonstrate our MSGField-based manipulation algorithm.

\subsection{Preliminaries: Gaussian Splatting}
\label{Preliminaries}
3DGS employs a set of 3D Gaussians as primitives to represent the scene's appearance and geometry. 
A 3D Gaussian primitive is parameterized as $\mathbf{g} \equiv (\mathbf{\boldsymbol{\mu}}, \mathbf{R}, \mathbf{s}, \mathbf{o}, \mathbf{c})$, where $\mathbf{\boldsymbol{\mu}} \in \mathbb{R}^3$ and $\mathbf{R} \in \mathbb{SO}(3)$ denote the 3D mean and orientation, while  $\mathbf{s} \in \mathbb{R}^3$, $\mathbf{o} \in \mathbb{R}$ and $\mathbf{c} \in \mathbb{R}^3$ correspond to scale, opacity, and color, respectively.  
A 2D Gaussian primitive can be regarded as a degenerate ellipsoid, retaining only the first two dimensions of $\mathbf{R}$ and $\mathbf{s}$,
For convenience, we use the same parameterization as for a 3D Gaussian, treating the first two dimensions of $\mathbf{R}$ as two principal tangential vectors $\mathbf{t}_u$ and $\mathbf{t}_v$ of 2D Gaussian.
They both employ $\alpha$-blending to render a pixel color $C$, as detailed in Eq.~\eqref{eq1}, 

\begin{equation}
    \label{eq1}
    C = \sum_{i \in \mathcal{N}} c_i \alpha_i \prod_{j=1}^{i-1}(1-\alpha_j),
\end{equation}
where $\mathcal{N}$ is ordered points overlapping the pixel, $c_i$ is the color of each point and $\alpha_i$ is the opacity multiplied Gaussian covariance.
Methods~\cite{qin2024langsplat,zhou2024feature3dgs,wangShapeMotion4D2024} replace $c_i$ in Eq.~\eqref{eq1} with each primitive's distance to the camera or its semantic features to generate a depth or semantic feature map.
Supervising the model with corresponding ground truth enables 3D reconstruction or semantic distillation; however, ensuring accuracy remains challenging.
For 3D reconstruction, the absence of distance constraints between Gaussian primitives can lead to $\alpha$-blended depths that fall between them.
In contrast, 2DGS offer a more accurate geometric representation and incorporate a depth distortion loss to regulate inter-primitive distances.
For feature distillation, $\alpha$-blending complicates maintaining meaningful semantic features, as blending colors like red and green to get yellow is valid, but blending different features is meaningless.
We will demonstrate this in our experiments in Section~\ref{Impact_of_Object-Centric}.

\subsection{Object-Centric Semantic Field Distillation}
\label{Semantic}
Based on the analysis in Section~\ref{Preliminaries}, we design a 
 field distillation framework to ensure feature consistency within objects, as shown in Fig.~\ref{pipeline}.
For preprocessing, YOLO-World~\cite{Cheng2024YOLOWorld} detects the object from multiple views using language prompts, and the detection results guide the Segment Anything Model (SAM)~\cite{SAM} to generate accurate object masks.
Next, we use ray-splat intersection method~\cite{weyrich2007ray-splat} to determine whether each primitive belongs to the object, a process we refer to as label rendering.
Specifically, for a primitive in the camera coordinate system,  we define the matrix $T$:
\begin{equation}
    T = \left(
            \begin{array}{ccc}
                \mathbf{t}_s & \mathbf{t}_v & \mathbf{\boldsymbol{\mu}} \\
                 0 & 0 & 1
            \end{array}
        \right)^T
        P^T \in \mathbb{R}^{3 \times 4},
\end{equation}
where $P \in \mathbb{R}^{4\times 4}$ is projection matrix of camera.
The projection of the 2D Gaussian onto image plane, with a center $\mathbf{p} = (p_1, p_2)$ and half extends of bounds $\mathbf{h} = (h_1, h_2)$, can be expressed as:
\begin{equation}
\begin{aligned}
    &p_i = d^{-1}(1,1,-1) (T_i \star T_4), \\
    &h_i = \sqrt{p_i^2 - d^{-1}(1,1,-1) (T_i \star T_i)},\\
    &d = (1,1,-1) (T_4 \star T_4),
\end{aligned}
\end{equation}
where $T_i$ is the $i$-th column of $T$, and $\star$ meas component-wise product.
The projection boundaries can be obtained as $x_1,x_2 = p_1 \pm h1$ and $y_1, y_2 = p_2 \pm h_2$, as illustrated  in Fig.~\ref{projection} a). 
A primitive is considered part of the object when its projection boundaries fall within the mask obtained from SAM, as shown in Fig.~\ref{projection} b).
However, background points within the view frustum also fall within the mask. 
To filter them out, we assign all primitives a black color and render the scene,  then compute the loss between the mask (where only the object remains white) and the rendered image. 
By backpropagating, only foreground points of the object receive gradients.
Points with gradients above a threshold are considered as target points and share the same feature $\mathbf{F}_{obj}$ extracted by CLIP.
To support 3D open-vocabulary segmentation, we follow~\cite{qin2024langsplat} to define the similarity $s$ between the text query $\mathbf{F}_{txt}$ and $\mathbf{F}_{obj}$ as: 
\begin{equation}
    \label{similarity}
   s =  min_i\frac{exp(\mathbf{F}_{txt}.\mathbf{F}_{obj})}{exp(\mathbf{F}_{txt}.\mathbf{F}_{obj}+exp(\mathbf{F}_{canon}^i.\mathbf{F}_{obj})},
\end{equation}
where $\mathbf{F}_{canon}^i$ is the CLIP embeddings of a predefined canonical phrase chosen from ``object'', ``things'', ``stuff'', and ``texture''.
The primitives with the highest similarity is segmented.
Given that the feature dimension is 512 or larger, each primitive is assigned a one-dimensional label attribute linked to object-level features to improve storage efficiency.
Thanks to the rapid rendering of 2DGS, the entire process, including preprocessing, takes less than 30 seconds.

\subsection{Motion Field Parameterization}
\label{Motion}
The Motion Field aims to recover primitive positions and orientations over time from a sequence of video frames $\{\mathbf{I}_{t_i} \in \mathbb{R}^{H\times W \times 3}\}_{i=0}^{t}$.
Specifically, for a 2D Gaussian primitive, the attributions at time $t_0$ are represented as $\mathbf{g}^{t_0} \equiv (\mathbf{\boldsymbol{\mu}}^{t_0}, \mathbf{q}^{t_0}, \mathbf{s}, \mathbf{o}, \mathbf{c})$. 
Here, we use the unit quaternion $\mathbf{q} \in \mathbb{S}^3$ to represent orientation for efficient optimization, as done in~\cite{kratimenos2023dynmf}.
Assuming that $\mathbf{s}$, $\mathbf{o}$ and $\mathbf{c}$ remain unchanged over time, the position and orientation at time $t_i$ are:
\begin{equation}
    \mathbf{\boldsymbol{\mu}}^{t_i} = \mathbf{\boldsymbol{\mu}}^{t_0} + \Delta\boldsymbol{\mu}_{0 \rightarrow t_i},
    \mathbf{q}^{t_i} = normalize(\mathbf{q}^{t_0} + \Delta \mathbf{q}_{0 \rightarrow t_i}),
\end{equation}
where $\Delta\boldsymbol{\mu}_{0 \rightarrow t_i}$ and $\Delta \mathbf{q}_{0 \rightarrow t_i}$ represent the motion changes we aim to capture.
We observe that the motion remaining consistent within a given area. 
In the extreme case of rigid body motion, all primitives share the same movement. 
Thus, optimizing each primitive’s motion individually is redundant and may lead to uncontrolled deformations.
Inspired by \cite{kratimenos2023dynmf}, we use a set of learnable motion bases $\{(\boldsymbol{\mu}^{(b)}_{0 \rightarrow t_i},\mathbf{q}^{(b)}_{0 \rightarrow t_i})\}_{b=1}^B$ to compactly represent scene motion. Each primitive's motion is decomposed into a combination of these bases, as expressed in Eq.~\eqref{motion_bases}, where $\{w^{(b)}\}_{b=1}^B$  represents the motion coefficient of the primitive.
\begin{equation}
\label{motion_bases}
    \Delta\boldsymbol{\mu}_{0 \rightarrow t_i} = \sum_{b=1}^{B} w^{(b)} \boldsymbol{\mu}^{(b)}_{0 \rightarrow t_i},
    \Delta \mathbf{q}_{0 \rightarrow t_i} = \sum_{b=1}^{B} w^{(b)} \mathbf{q}^{(b)}_{0 \rightarrow t_i}.
\end{equation}
We focus on optimizing the dynamic primitives, treating other as static. 
To accelerate optimization, we employ sparse frame sampling to $\mathbf{I}_{t_i}$.
However, this leads to sampled frames with minimal overlap, causing most primitives to compute losses with the background and hindering convergence.
To address this, we introduce Dice loss $L_{Dice}$, which emphasizes global pixel distribution over pixel-wise RGB or SSIM losses~\cite{huang2DGaussianSplatting2024}.
The $L_{Dice}$ is:
\begin{equation}
    L_{Dice} = 1 - Dice(I_{opacity}, I_{mask}).
\end{equation}
Here, $I_{opacity}$ represents the image rendered with the dynamic primitives' opacity, and $I_{mask}$ is the dynamic object mask for the current frame.
This loss provides a global optimization direction, allowing the primitives to quickly adjust to the region corresponding to the current frame.
We retained the rigidity loss from~\cite{kratimenos2023dynmf} to encourage rigidity between primitives.
Our total loss $L$ is:
\begin{equation}
    L = L_{RGB} + \lambda_1 L_{SSIM} + \lambda_2 L_{Dice} + \lambda_3L_{Rigidity}.
\end{equation}
For rigid body motion, where all primitives share the same $\mathbb{SE}(3)$ transformation, we assign each primitive the same fixed motion coefficient and only optimize the motion bases.
Rotation and translation changes are applied as an $\mathbb{SE}(3)$ matrix to the original primitives.

\begin{algorithm}[t]
\caption{Robot Manipulation with MSGField}
\begin{algorithmic}[1]
\label{algorithm1}
\STATE \textbf{Initial}: \textbf{\emph{robot}}.move($P_{home}$), \textbf{\emph{robot}}.grasp\_open
\STATE \textbf{Input}: $L_{operation}, L_{obj} \leftarrow language$
\STATE $G_{obj} \leftarrow \text{\textbf{\emph{MSGField}}.semantic}(L_{obj})$
\STATE $P_{obj} \leftarrow \text{\textbf{\emph{MSGField}}.geometry}(G_{obj})$
\WHILE{NOT (\textbf{\emph{robot}}.current\_pose == $P_{obj}$)}
    \STATE $I \leftarrow$ Camera.perception
    \IF{detect\_motion($I$)}
        \STATE \textbf{\emph{robot}}.stop
        \STATE $M_{obj} \leftarrow \text{\textbf{\emph{MSGField}}.motion}(G_{obj}, I)$
        \STATE \textbf{\emph{MSGField}}.update($M_{obj}$)
        \STATE $P_{obj} \leftarrow \text{\textbf{\emph{MSGField}}.geometry}(G_{obj})$
    \ELSE
        \STATE \textbf{\emph{robot}}.move($P_{obj}$)
    \ENDIF
\ENDWHILE
\IF{$L_{operation}$ == `grasp'}
    \STATE \textbf{\emph{robot}}.grasp\_close
\ELSIF{$L_{operation}$ == `place'}
    \STATE \textbf{\emph{robot}}.grasp\_open
\ELSE
    \STATE \textbf{Raise} NotImplementedError
\ENDIF
\end{algorithmic}
\end{algorithm}

\subsection{MSGFiled-Based Language-Guided Manipulation}
\label{Grasp}
\begin{figure}[t]
\vspace{0.3cm}
\centering
 \begin{minipage}{1\linewidth}
    \centering
        \includegraphics[width=1\linewidth]{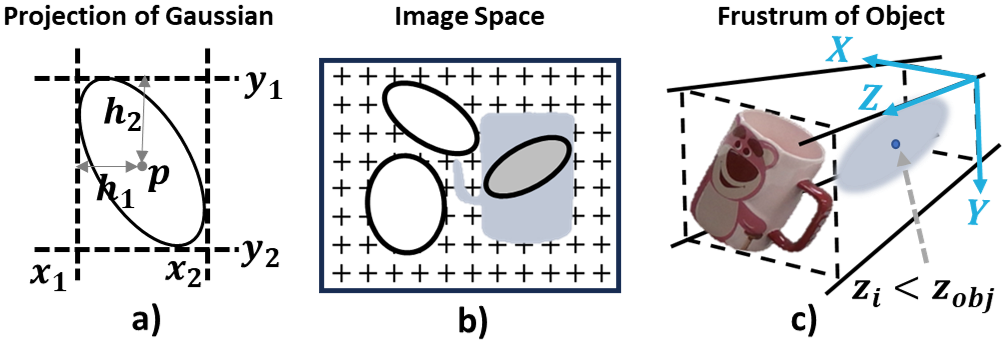}
    \end{minipage}
\caption{a). Projection of the 2D Gaussian on the image.
b). Determining whether a Gaussian primitive falls within the mask. The gray primitive is inside.
c). The gray primitives are the ones occluding the object.}
\label{projection}
\vspace{-1.5em}
\end{figure}

Leveraging the precise motion, semantic, and geometric information provided by MSGField, robots can accurately manipulate objects in dynamic real-world environments based on natural language instructions.
The pseudocode of our language-guided manipulation algorithm is illustrated in Algorithm~\ref{algorithm1}.
We perform both grasping and placing operations, primarily by controlling the end-effector's pose and the gripper's opening and closing using the robot API.
Specifically, the object primitives $G_{obj}$ are segmented according to the text query $L_{obj}$, and the end-effector's pose $P_{obj}$ is determined based on~\cite{fang2020graspnet} (Line 3-4 in Algorithm~\ref{algorithm1}).
We convert the precise mesh from MSGField into a pseudo RGB-D point cloud, which serves as the input for~\cite{fang2020graspnet}. 
We use a straightforward feature matching method to detect object motion.
If the object moves, the MSGField will update, allowing the robot to adjust and execute the manipulation effectively (Line 5-15 in Algorithm~\ref{algorithm1}).
We observed that~\cite{fang2020graspnet} is sensitive to variations in camera coordinate systems, and a single camera perspective often encounters occlusion issues.
Therefor, we sample multiple camera perspectives that minimize occlusion and combine the grasps derived from them.
Specifically, for a camera view with a projection matrix $P$, we apply perspective division transform the positions of primitives $\boldsymbol{\mu_i}$ into Normalized Device Coordinates (NDC) space:
\begin{equation}
    \boldsymbol{\mu_i}^{ndc} = \boldsymbol{\mu_i}^{clip}[:3] / \boldsymbol{\mu_i}^{clip}[3],
    \quad
    \boldsymbol{\mu_i}^{clip} = P \left(
    \begin{array}{c}
         \boldsymbol{\mu_i} \\
         1 
    \end{array}
    \right).
\end{equation}
The viewing frustum transforms into a standard cuboid in NDC space, where occlusion points can be easily identified, as shown in Fig.\ref{projection} c). 
We select views with minimal occlusion and filter out grasps misaligned with the robotic arm's orientation to obtain more robust grasp poses.

%% file: experiments.tex
\section{EXPRTIMENTS}
\subsection{Setup, Dataset and Metric}
\subsubsection{Setup}
Fig.~\ref{msg_field}  illustrates our experimental setup. We use a Kinova-Gen3 robotic arm equipped with a Robotiq 2F-85 gripper. 
Two RealSense-D455 depth cameras are positioned on either side to capture object motion. 
Notably, our method relies solely on RGB data.
Both cameras are precisely calibrated using hand-eye calibration~\cite{handeye_calib} to ensure alignment with the robot arm's base.
Another camera of the same type, along with the two cameras, provides panoramic views. 
The optimization runs on an RTX 3090 remote server, while the laptop is used to control the robot and transfer messages.
We integrate and manage the hardware components with ROS.



\subsubsection{Dataset}
Our dataset includes both static and dynamic scenes. 
The static scenes are categorized into three groups based on object type: daily objects, small objects, and flexible objects, and the dynamic scenes are divided into rigid and non-rigid body motion.
Each category contains two distinct scenarios, with the entire dataset spanning up to 30 distinct object types, as shown in Fig.~\ref{Scenes}.
For each scene, we capture 16 panoramic images and use COLMAP~\cite{sfm} to estimate the extrinsic parameters, which were then scaled and aligned to the robot arm's base with extrinsic parameters of the two pre-calibrated cameras. 
For dynamic scenes, we capture an additional video sequence using two calibrated cameras.

\subsubsection{Metric}
We assess the methods based on the operational success rate, consistent with~\cite{shenDistilledFeatureFields2023, fang2020graspnet}. 
For each scene, we conduct a total of 30 manipulations attempts.
A successful grasp occurs when the gripper picks up the specified object without external intervention and holds it suspended. 
A successful placement requires the robot to place the object within the designated area without disturbing the existing arrangement.

\subsection{Impliment Details}
\label{Impliment}
We first use 2D Gaussian Splatting to train a geometry field. We train it for 20k iterations in our experiments (10k iterations is enough for most scene). 
The feature field and motion field are generated on-site. 
For the feature field, we use OpenCLIP ViT-B/16 model and SAM ViT-H model to extract features.
The motion field is optimized using the Adam optimizer, with total 300 iterations for rigid body motion and 1000 iterations for non-rigid motion per frame. 
The number of motion bases is set to 10.
We choose F3RM~\cite{shenDistilledFeatureFields2023}, a recent work which employs NeRF to distill a feature field enabling language-guided manipulation, as our baseline.
Following the setting in~\cite{shenDistilledFeatureFields2023}, we collect 40 panoramic images and optimize each scene for 2,000 iterations for it. Since F3RM cannot handle dynamic scenes,  we directly compare our method with point cloud-based approach GraspNet~\cite{fang2020graspnet} to assess the accuracy of the motion field. 
The real-time calibrated RGB-D frames are used as input for~\cite{fang2020graspnet}, with the target object for grasping manually specified.

\begin{table}[t]
\vspace{1em}
    \setlength{\abovecaptionskip}{0em}
    \centering
    \caption{Language-Guided Manipulation Success Rate}
    \label{table:Total_Scene}
    \resizebox{\linewidth}{!}{
    \begin{tabular}{*{4}{c}}
        \toprule
        \multirow{2}*{Scene} & \multicolumn{3}{c}{Success Rate($\%$) $\uparrow$} \\
        \cline{2-4}
         & F3RM~\cite{shenDistilledFeatureFields2023} & GraspNet(RGB-D)~\cite{fang2020graspnet} & MSGField\\
         \midrule
         Daily Objects & 76.7 & - & \textbf{90.0} \\
         Small Objects & 30.0 & - & \textbf{56.7} \\
         Flexible Objects & 23.3 & - & \textbf{80.0} \\
         Place & 66.7 & - & \textbf{90.0} \\
         \midrule
         Rigid Body Motion$^\dagger$ & - & \textbf{86.7} & 83.3 \\
         Non-Rigid Motion$^\dagger$ & - & \textbf{90.0} & 43.3 \\
         \midrule
         Static Avg. & 49.2 & - & \textbf{79.2} \\
         Dynamic Avg. & - & \textbf{88.4} & 63.3 \\
         \bottomrule
         \multicolumn{2}{l}{$^\dagger$ Without language guide.}
    \end{tabular}}
\vspace{-2em}
\end{table}

\begin{figure}[t]
\vspace{0.3cm}
    \begin{minipage}{1\linewidth}
    \centering
        \includegraphics[width=0.49\linewidth]{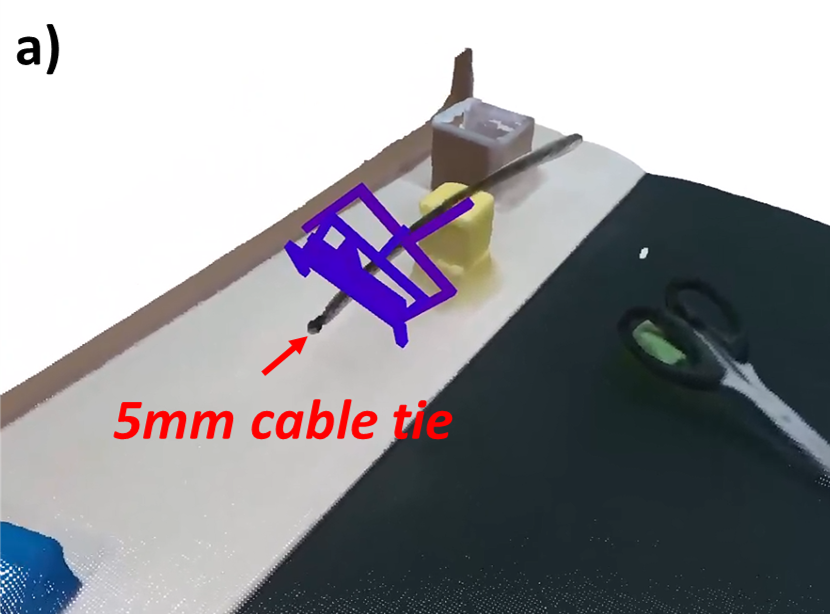}
        \includegraphics[width=0.49\linewidth]{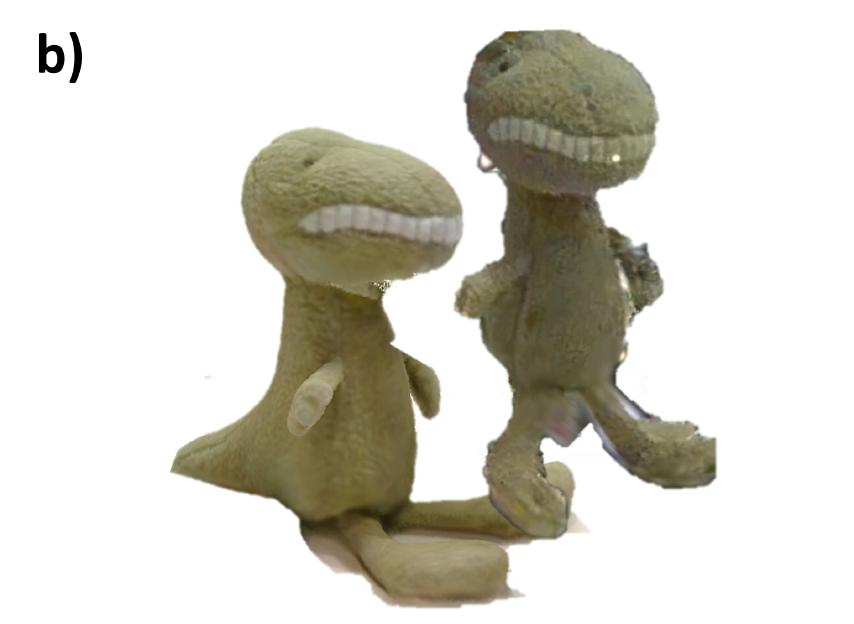}
    \end{minipage}
    \caption{a) MSGField produces accurate meshes and can  generate robust  grasp poses, even for a 5mm cable tie. b) MSGField effectively optimizes non-rigid motion, recovering a toy's transition from sitting to standing.}
    \label{details}
\vspace{-1.0em}
\end{figure}


\begin{table}[t]
\vspace{1em}
\tiny
    \setlength{\abovecaptionskip}{0em}
    \centering
    \caption{Static Scene Grasping Without Language Guiding}
    \label{table:Grasping_Without_Language_Guiding}
    \resizebox{\linewidth}{!}{
    \begin{tabular}{*{3}{c}}
        \toprule
        \multirow{2}*{Scene} & \multicolumn{2}{c}{Success Rate($\%$) $\uparrow$} \\
        \cline{2-3}
         & GraspNet(RGB-D based)~\cite{fang2020graspnet} & MSGField\\
         \midrule
         Daily Objects & \textbf{100} & 93.3 \\
         Small Objects & 50.0 & \textbf{90.0} \\
         Flexible Objects & \textbf{90.0} & 86.7 \\
         \midrule
         All Avg.& 80.0 & \textbf{90.0} \\
         \bottomrule
    \end{tabular}}
\vspace{-2em}
\end{table}

\begin{figure*}[t]
\vspace{0.3cm}
    \begin{minipage}{1\linewidth}
    \centering
        \includegraphics[width=0.19\linewidth]{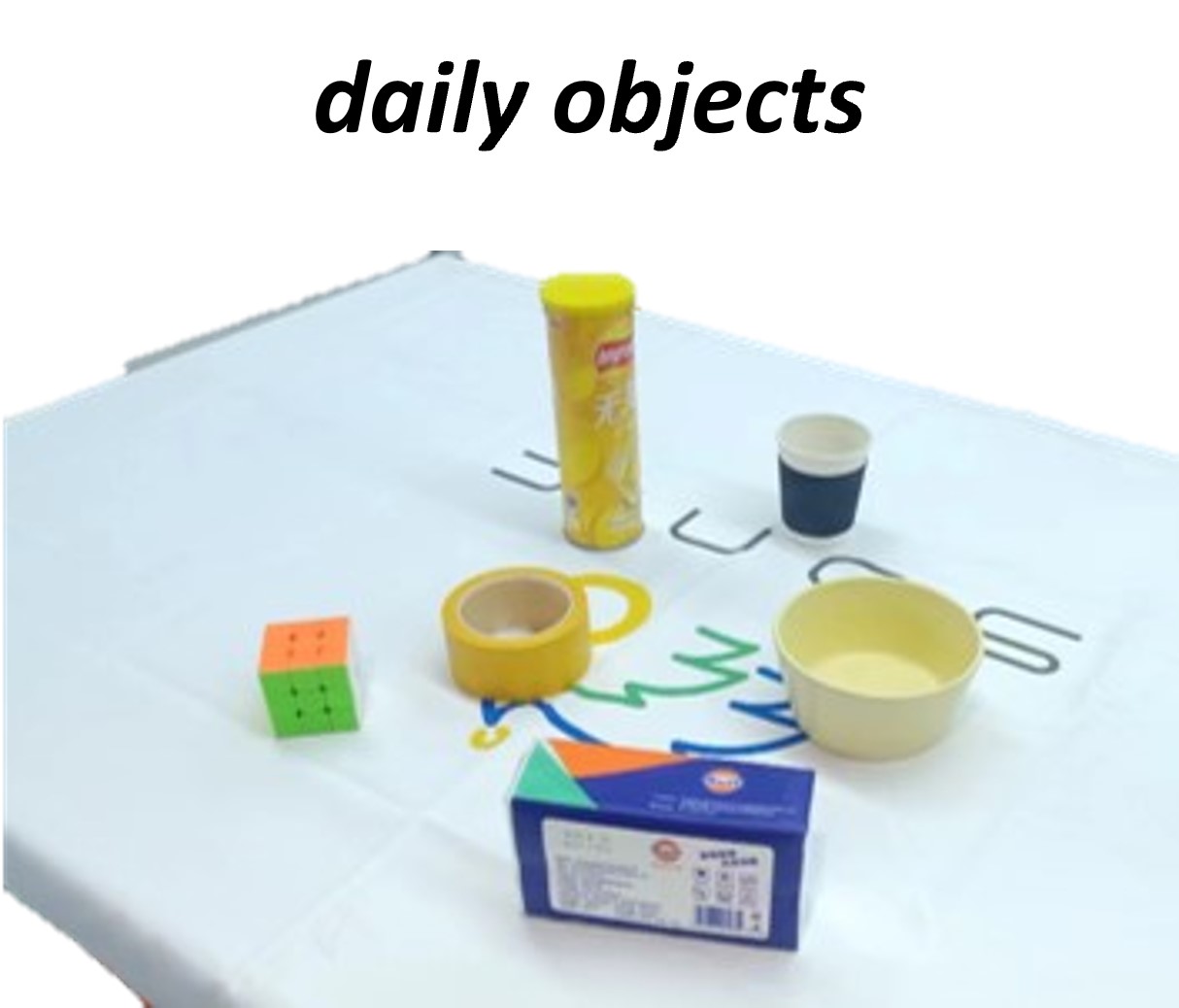}
        \includegraphics[width=0.19\linewidth]{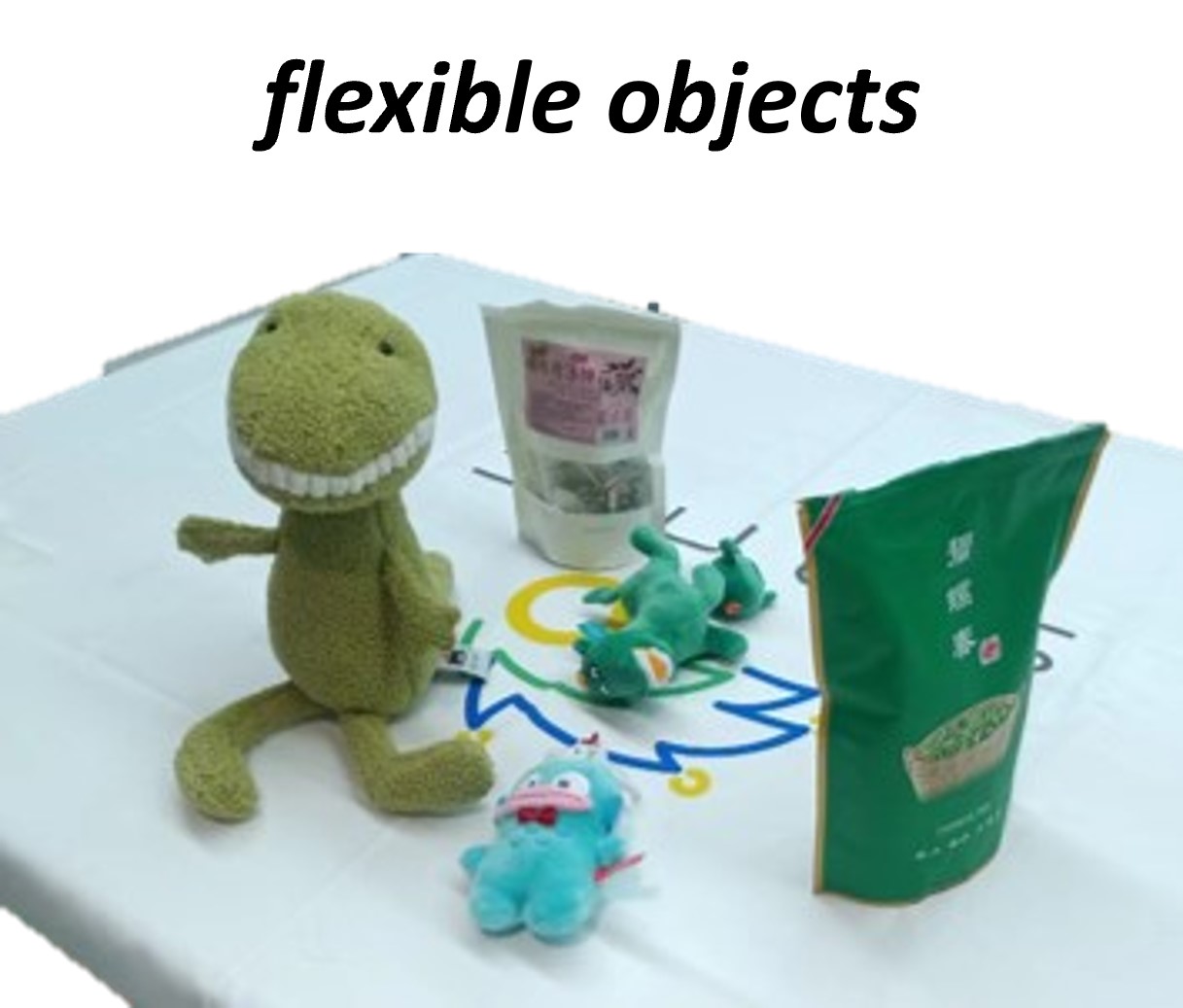}
        \includegraphics[width=0.19\linewidth]{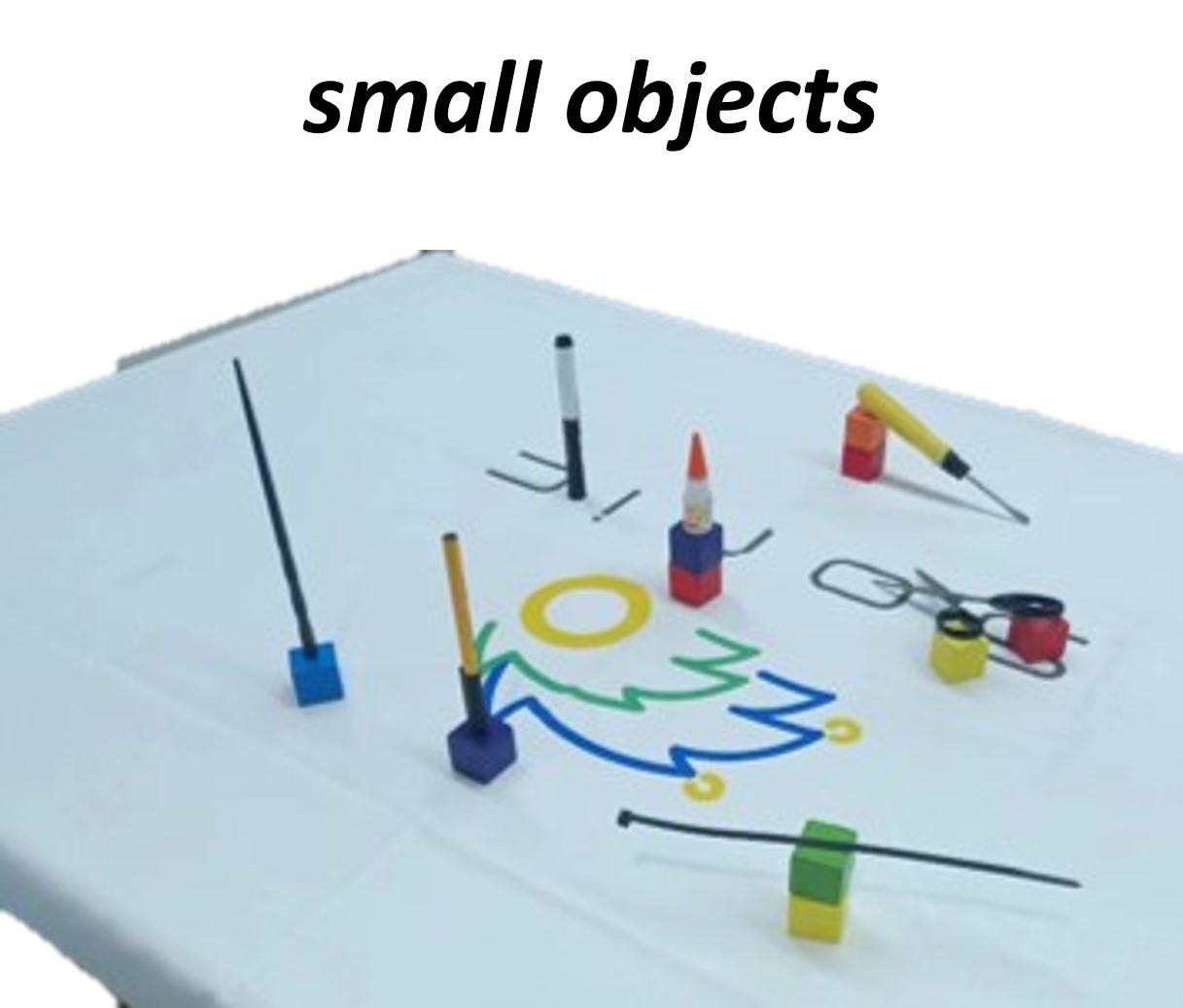}
        \includegraphics[width=0.19\linewidth]{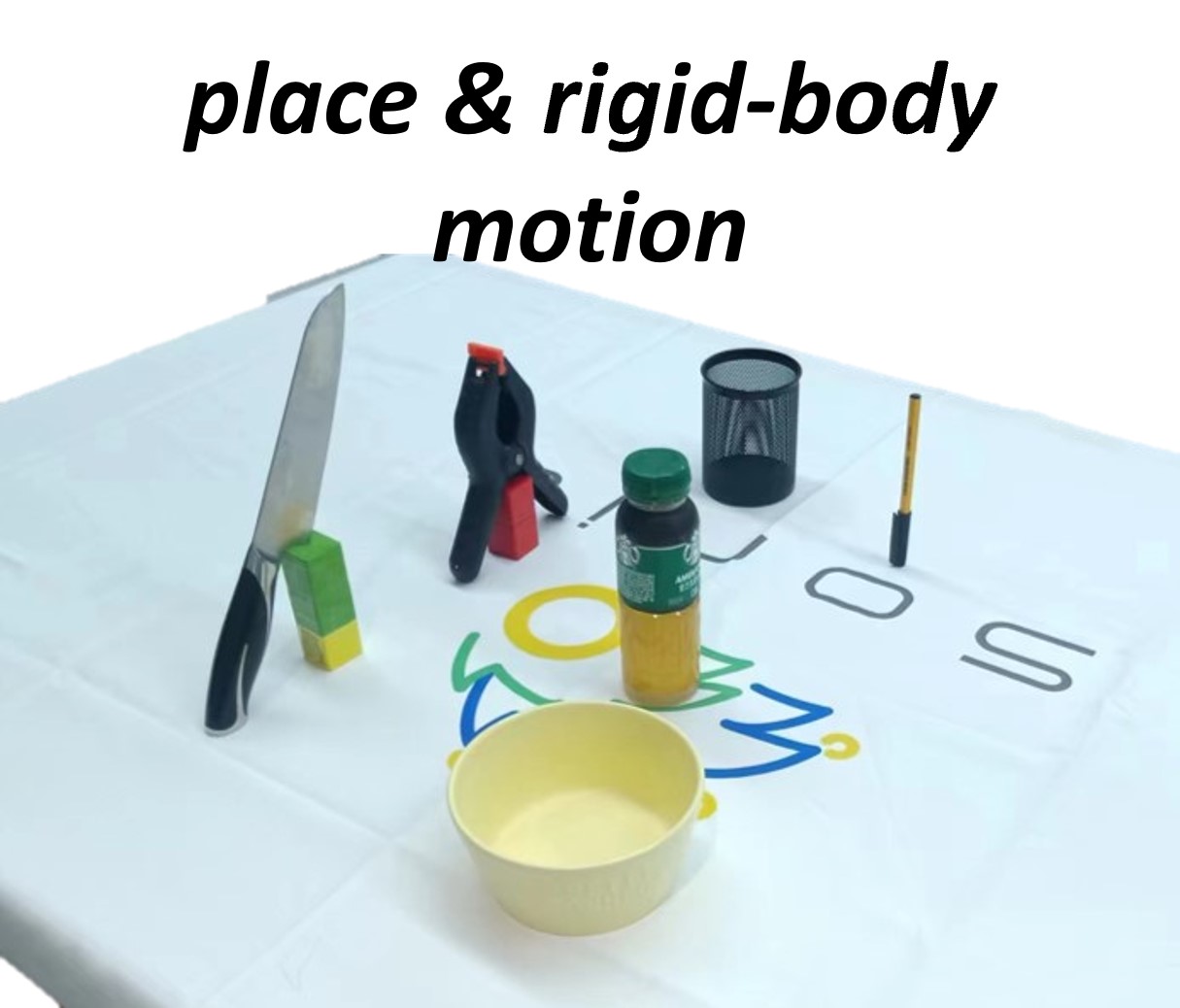}
        \includegraphics[width=0.19\linewidth]{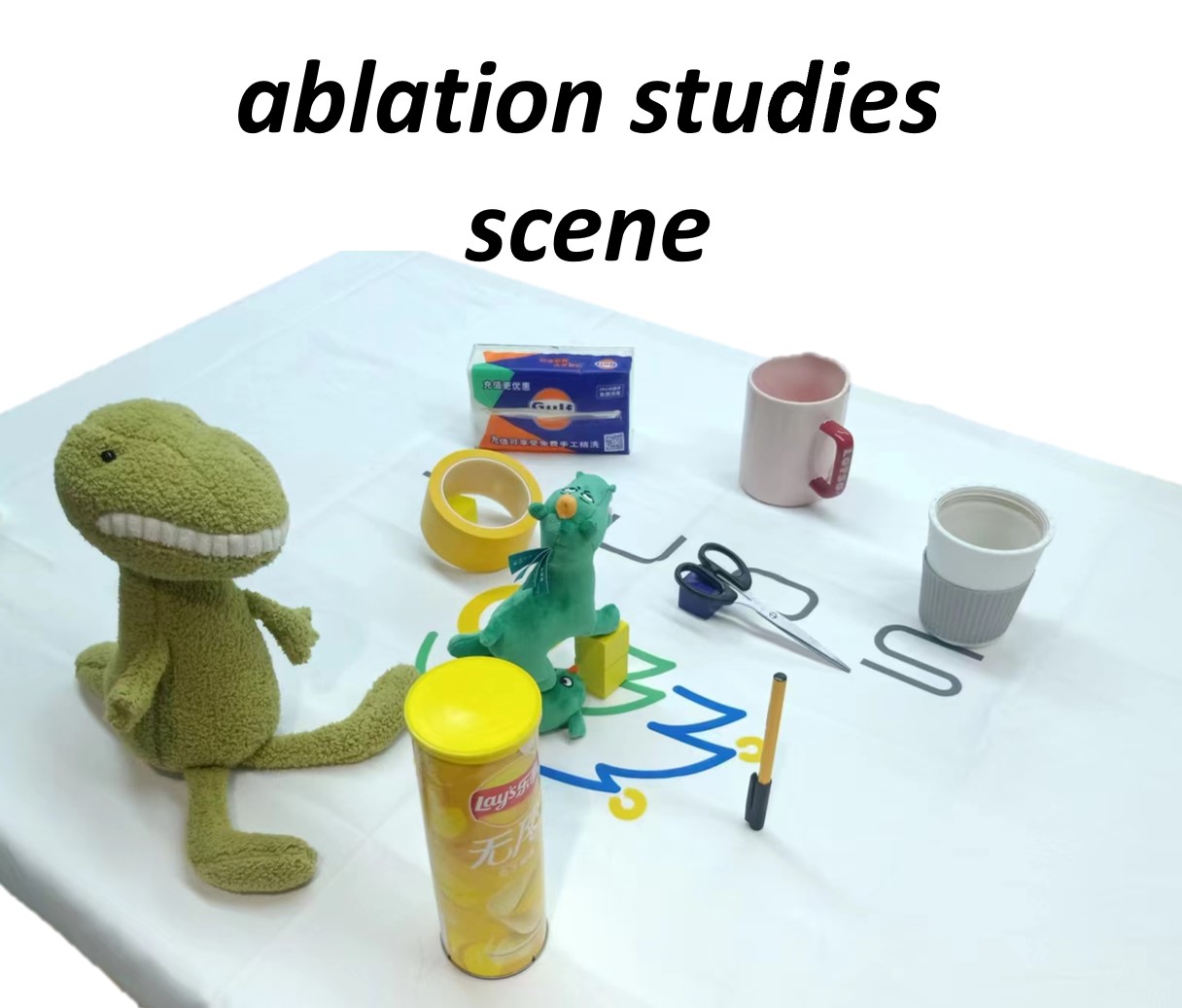}
    \end{minipage}
    \caption{Visualization of certain scenes from the dataset. We strongly recommend watching the supplementary video for more detailed information.}
    \label{Scenes}
\vspace{-1.0em}
\end{figure*}

\subsection{Overall Results and Analysis}
\vspace{2pt}\noindent\textbf{Static Scene.}
As shown in Table~\ref{table:Total_Scene}, our method achieved an average success rate of 79.2$\%$ in static scenes, surpassing~\cite{shenDistilledFeatureFields2023} by a large margin, demonstrating its accuracy and generalization capability.
\cite{shenDistilledFeatureFields2023} performs well with daily objects but struggles with the small and flexible object. 
This is mainly due to insufficient modeling accuracy and the limitations of its matching-based grasp detection algorithm, which struggles to generalize to flexible items with complex surfaces. 
Our method's success rate also significantly declines with small objects, primarily because their indistinct features make natural language matching difficult. 
For example, ``black cable tie'' is frequently mismatched, which is a major source of fail in our approach.
The main reason we outperform~\cite{shenDistilledFeatureFields2023} in placement tasks is that querying with ``black pen container'' or ``yellow bowl'' in~\cite{shenDistilledFeatureFields2023} often includes many desktop points, leading to inaccurate position estimation and failure.
In contrast, our object-centered semantic field effectively isolates the queried objects.

\begin{figure}[t]
\vspace{0.3cm}
    \begin{minipage}{1\linewidth}
    \centering
        \includegraphics[width=0.49\linewidth]{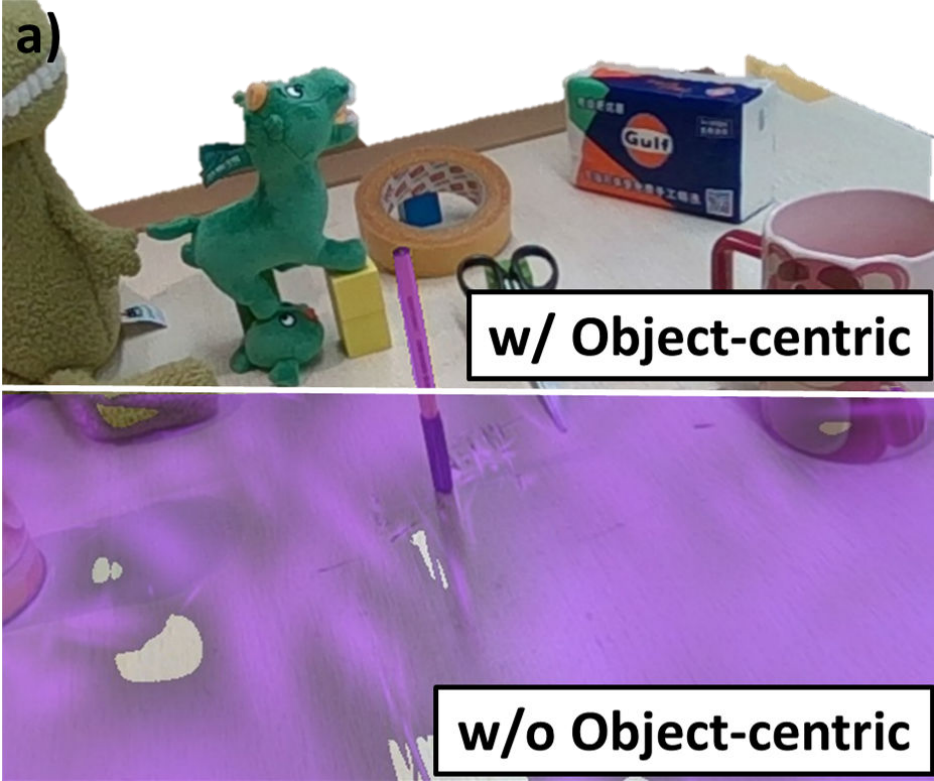}
        \includegraphics[width=0.49\linewidth]{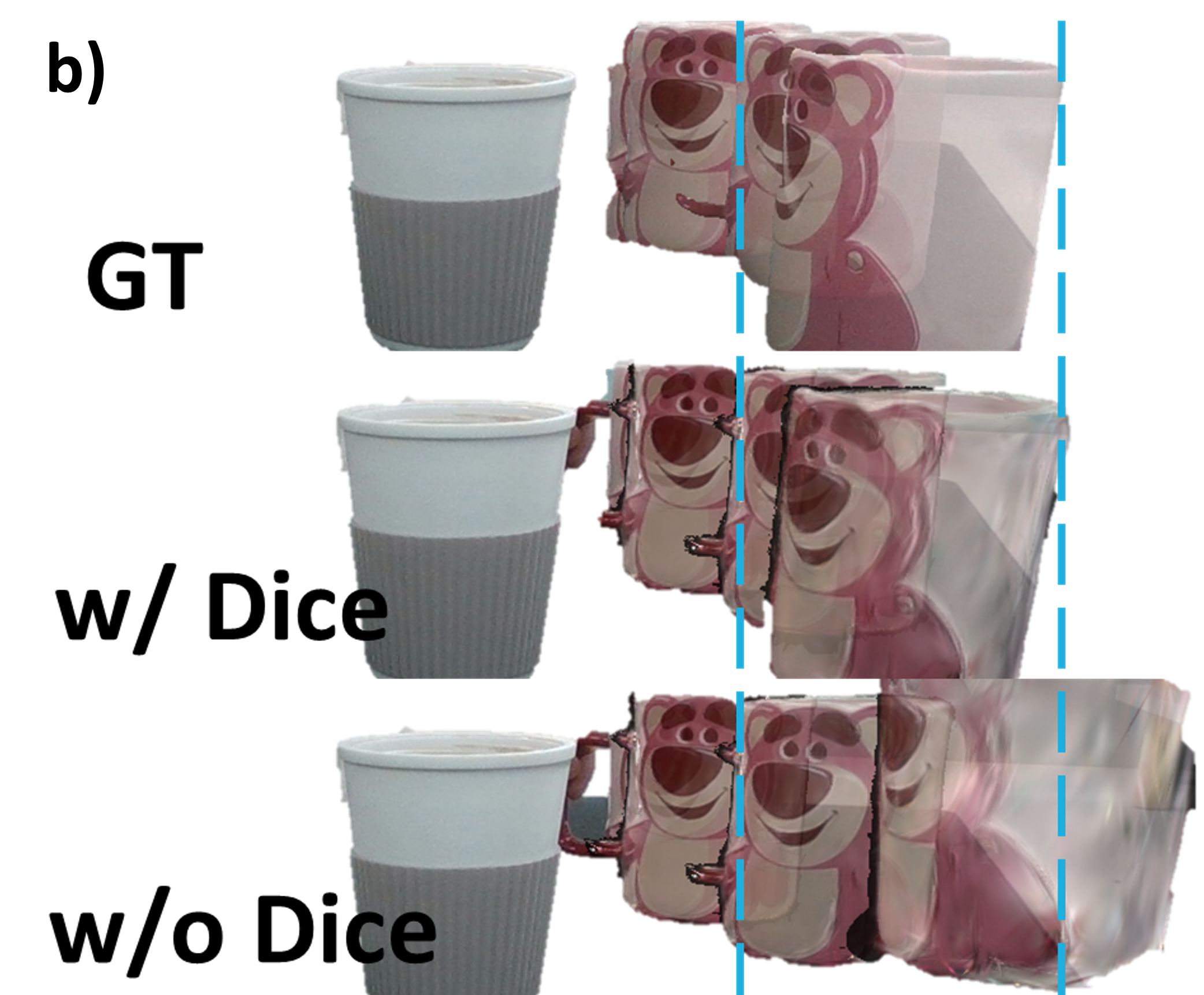}
    \end{minipage}
    \caption{a) Use a ``yellow pen'' to query and project the semantic field onto the image. Object-centric feature distillation focuses on the corresponding primitives as the purple visualization results.
b) Without the constraint of the Dice loss, the motion fails to converge.}
    \label{Ablation_Study}
\vspace{-2em}
\end{figure}

\vspace{2pt}\noindent\textbf{Dynamic Scene.}
In dynamic scenes, our method is nearly on par with GraspNet for rigid motion, without the requirement for a depth sensor. 
When the object moves farther from the camera or to the camera's side, the accuracy of~\cite{fang2020graspnet} begins to decline, whereas our method does not encounter this issue.
Our method often fails when an object reveals previously unseen sides, such as a side contacted with table.
For non-rigid motion, the motion field is well-optimized in camera-supervised views within 2 minutes, as shown in Fig.~\ref{details} b), resulting in a success rate of 43.3$\%$. 
The success rate drops as the limited supervision from only two viewpoints makes it difficult to optimize the motion of primitives in unsupervised views.
~\cite{fang2020graspnet} directly uses real-time RGB-D as inputs, it can achieve the same  performance as in static scenes.

\subsection{Reconstruction Accuracy Analysis}
To further assess our reconstruction accuracy, we compared our method with GraspNet~\cite{fang2020graspnet}.
We convert the geometric information from MSGField to match the same perspective as GraspNet inputs and manually specify the objects to be grasped.
As shown in Table~\ref{table:Grasping_Without_Language_Guiding}, our method achieve much higher accuracy than~\cite{fang2020graspnet} for small objects.
The main challenge with RGB-D point clouds is their difficulty in accurately modeling small objects, especially as distance increases. 
In contrast, the MSGField provides highly precise geometry, clearly capturing objects as small as a 5mm cable tie, as shown in Fig.~\ref{details} a).
For regular-sized and flexible objects, our method's modeling accuracy is on par with RGB-D.
The slight difference in success rate compared to~\cite{fang2020graspnet} is primarily due to minor distortions in some scenes caused by low lighting conditions.

\subsection{Ablation Studies}
In this section, we demonstrate the effectiveness and efficiency of different components of our method in terms of grasping success rate and optimization time. 
As shown in Fig.~\ref{Scenes}, we designed a scene with various objects, moving the pink mug for simulating rigid body motion and the green dinosaur toy for non-rigid motion scenarios.

\vspace{2pt}\noindent\textbf{Impact of 2D Gaussian.}
We replaced the geometric component of the scene representation with 3D Gaussian, as done in GaussianGrasper~\cite{zhengGaussianGrasper3DLanguage2024}.
To eliminate the effects of other factors, we manually specified the target object for grasping in a static environment. 
As shown in Table~\ref{table:Ablation}, using 3D Gaussian  results in a 63.4$\%$ decrease in performance. 
Visualization of the reconstruction results reveals significant surface distortions with 3D Gaussians, causing the generated grasp to shift, leading to failed grasps.
While 2D Gaussian has a longer optimization time per iteration, it achieves much better surface reconstruction quality.

\vspace{2pt}\noindent\textbf{Impact of Object-Centric Feature Distillation.}
\label{Impact_of_Object-Centric}
To validate the effectiveness of Object-Centric feature distillation, we replaced our method with LangSplat~\cite{qin2024langsplat}, a recent per-primitive feature distillation approach. 
Following the setup in~\cite{qin2024langsplat}, we use the CLIP and SAM models as mentioned in Section.~\ref{Impliment} to extract image features and compress the CLIP features to 3 dimensions with autoencoder. 
The feature field was trained for 30,000 iterations.
~\cite{qin2024langsplat} achieves excellent semantic segmentation performance on 2D images, consistent with the results reported in the paper.
However, for 3D segmentation, using the same decoder to recover features to 512 dimensions, it becomes nearly impossible to directly query the corresponding primitives using natural language, as shown in Fig.~\ref{Ablation_Study} a). 
Even increasing the compressed feature dimensions (up to 32 dimensions on RTX 3090) to reduce information loss yielded unsatisfactory results.
This supports our analysis in Section~\ref{Preliminaries}, where we showed that features lose their meaning after $\alpha$-blending.
And the time consumption (without preprocessing) is also unacceptable.

\begin{table}[t]
\vspace{1em}
    \setlength{\abovecaptionskip}{0em}
    \centering
    \caption{Ablation Results About Success Rate}
    \label{table:Ablation}
    \resizebox{\columnwidth}{!}{
    \begin{tabular}{*{5}{c}}
        \toprule
        \multicolumn{3}{c}{Component} & \multicolumn{2}{c}{Success Rate($\%$) $\uparrow$} \\
        \midrule
        2D-GS & Object-Centric &  Dice Loss & Static Obj. & Dynamic Obj. \\
        \midrule
        \ding{52}& \ding{52} & \ding{52} & \textbf{86.7} & \textbf{56.7} \\
        & \ding{52} & \ding{52} & 23.3 & - \\
        \ding{52} &  & \ding{52} & 0.0 & 0.0 \\
        \ding{52} & \ding{52} &  & - & 13.3 \\
        \bottomrule
    \end{tabular}}
\vspace{-2em}
\end{table}

\begin{table}[t]
\vspace{1em}
    \setlength{\abovecaptionskip}{0em}
    \centering
    \caption{Ablation Results About Training Time}
    \label{table:Ablation}
    \resizebox{\columnwidth}{!}{
    \begin{tabular}{*{6}{c}}
        \toprule
        \multicolumn{3}{c}{Component} & \multicolumn{3}{c}{Traing Time(min)$\downarrow$} \\
        \midrule
        2D-GS & Object-Centric &  Dice Loss & Geomerty & Semantic & Motion \\
        \midrule
        \ding{52}& \ding{52} & \ding{52} & 20.6 & \textbf{0.5} & \textbf{0.1} \\
        & \ding{52} & \ding{52} & \textbf{11.3} & 0.5 & 0.1 \\
        \ding{52} &  & \ding{52} & 20.6 & 16.9 & 0.1 \\
        \ding{52} & \ding{52} &  &  20.6 & 0.5 & 0.1\\
        \bottomrule
    \end{tabular}}
\vspace{-2em}
\end{table}

\vspace{2pt}\noindent\textbf{Impact of Dice Loss.}
To assess the effectiveness of Dice loss, we remove it and optimize the object's motion. 
As shown in Fig.~\ref{Ablation_Study} b), optimization can be converged with minor translations without Dice loss. 
However, with rotations or larger movements, relying solely on per-pixel loss fails to converge, as even slight movements can significantly affect RGB loss.
Integrating  Dice loss improves motion estimation, ensuring better position optimization and a 43.4$\%$ increase in manipulation success rate.


%% file: conclusions.tex
\section{CONCLUSIONS}
This paper presents a unified scene representation that integrates motion, semantic, and geometric information. 
It effectively captures object-level semantics and spatial motion, and its compact nature allows rapid optimization. 
However, the method has its limitations. 
Firstly, incorporating 2D Gaussian Splatting results in higher computational costs. 
Moreover, under supervision by only two RGB cameras, it has difficulty precisely optimizing highly complex non-rigid motions.
Nevertheless, advancements in robotics and compute vision are expected to address these challenges. 
Moving forward, we aim to enhance reconstruction speed and explore the application of this system in mobile robots.

%% file: root.bbl
\begin{thebibliography}{10}
\providecommand{\url}[1]{#1}
\csname url@rmstyle\endcsname
\providecommand{\newblock}{\relax}
\providecommand{\bibinfo}[2]{#2}
\providecommand\BIBentrySTDinterwordspacing{\spaceskip=0pt\relax}
\providecommand\BIBentryALTinterwordstretchfactor{4}
\providecommand\BIBentryALTinterwordspacing{\spaceskip=\fontdimen2\font plus
\BIBentryALTinterwordstretchfactor\fontdimen3\font minus \fontdimen4\font\relax}
\providecommand\BIBforeignlanguage[2]{{%
\expandafter\ifx\csname l@#1\endcsname\relax
\typeout{** WARNING: IEEEtran.bst: No hyphenation pattern has been}%
\typeout{** loaded for the language `#1'. Using the pattern for}%
\typeout{** the default language instead.}%
\else
\language=\csname l@#1\endcsname
\fi
#2}}

\bibitem{wangSparseDFFSparseViewFeature2024}
Q.~Wang, H.~Zhang, C.~Deng, Y.~You, H.~Dong, Y.~Zhu, and L.~Guibas, ``Sparsedff: Sparse-view feature distillation for one-shot dexterous manipulation,'' \emph{arXiv preprint arXiv:2310.16838}, 2023.

\bibitem{zhengGaussianGrasper3DLanguage2024}
Y.~Zheng, X.~Chen, Y.~Zheng, S.~Gu, R.~Yang, B.~Jin, P.~Li, C.~Zhong, Z.~Wang, L.~Liu, \emph{et~al.}, ``Gaussiangrasper: 3d language gaussian splatting for open-vocabulary robotic grasping,'' \emph{arXiv preprint arXiv:2403.09637}, 2024.

\bibitem{wangFieldsDynamic3D2023}
Y.~Wang, Z.~Li, M.~Zhang, K.~Driggs-Campbell, J.~Wu, L.~Fei-Fei, and Y.~Li, ``D3fields: Dynamic 3d descriptor fields for zero-shot generalizable robotic manipulation,'' \emph{arXiv preprint arXiv:2309.16118}, 2023.

\bibitem{kerbl3DGaussianSplatting2023}
B.~Kerbl, G.~Kopanas, T.~Leimk{\"u}hler, and G.~Drettakis, ``3d gaussian splatting for real-time radiance field rendering.'' \emph{ACM Trans. Graph.}, vol.~42, no.~4, pp. 139--1, 2023.

\bibitem{mildenhallNeRFRepresentingScenes2020}
B.~Mildenhall, P.~P. Srinivasan, M.~Tancik, J.~T. Barron, R.~Ramamoorthi, and R.~Ng, ``Nerf: Representing scenes as neural radiance fields for view synthesis,'' \emph{Communications of the ACM}, vol.~65, no.~1, pp. 99--106, 2021.

\bibitem{shenDistilledFeatureFields2023}
W.~Shen, G.~Yang, A.~Yu, J.~Wong, L.~P. Kaelbling, and P.~Isola, ``Distilled feature fields enable few-shot language-guided manipulation,'' \emph{arXiv preprint arXiv:2308.07931}, 2023.

\bibitem{huang2DGaussianSplatting2024}
B.~Huang, Z.~Yu, A.~Chen, A.~Geiger, and S.~Gao, ``2d gaussian splatting for geometrically accurate radiance fields,'' in \emph{ACM SIGGRAPH 2024 Conference Papers}, 2024, pp. 1--11.

\bibitem{jauhriLearningAnyView6DoF2024}
S.~Jauhri, I.~Lunawat, and G.~Chalvatzaki, ``Learning any-view 6dof robotic grasping in cluttered scenes via neural surface rendering,'' \emph{arXiv preprint arXiv:2306.07392}, 2023.

\bibitem{object-centric-3D-features}
G.~Tziafas, Y.~Xu, Z.~Li, and H.~Kasaei, ``3d feature distillation with object-centric priors,'' \emph{arXiv preprint arXiv:2406.18742}, 2024.

\bibitem{lerf-togo}
A.~Rashid, S.~Sharma, C.~M. Kim, J.~Kerr, L.~Y. Chen, A.~Kanazawa, and K.~Goldberg, ``Language embedded radiance fields for zero-shot task-oriented grasping,'' in \emph{7th Annual Conference on Robot Learning}, 2023.

\bibitem{GeFF}
R.-Z. Qiu, Y.~Hu, G.~Yang, Y.~Song, Y.~Fu, J.~Ye, J.~Mu, R.~Yang, N.~Atanasov, S.~Scherer, \emph{et~al.}, ``Learning generalizable feature fields for mobile manipulation,'' \emph{arXiv preprint arXiv:2403.07563}, 2024.

\bibitem{tziafas20243dfd}
G.~Tziafas, Y.~Xu, Z.~Li, and H.~Kasaei, ``3d feature distillation with object-centric priors,'' \emph{arXiv preprint arXiv:2406.18742}, 2024.

\bibitem{yu2021pixelnerf}
A.~Yu, V.~Ye, M.~Tancik, and A.~Kanazawa, ``pixelnerf: Neural radiance fields from one or few images,'' in \emph{Proceedings of the IEEE/CVF conference on computer vision and pattern recognition}, 2021, pp. 4578--4587.

\bibitem{bozic2020deepdeform}
A.~Bozic, M.~Zollhofer, C.~Theobalt, and M.~Nie{\ss}ner, ``Deepdeform: Learning non-rigid rgb-d reconstruction with semi-supervised data,'' in \emph{Proceedings of the IEEE/CVF Conference on Computer Vision and Pattern Recognition}, 2020, pp. 7002--7012.

\bibitem{dou2016fusion4d}
M.~Dou, S.~Khamis, Y.~Degtyarev, P.~Davidson, S.~R. Fanello, A.~Kowdle, S.~O. Escolano, C.~Rhemann, D.~Kim, J.~Taylor, \emph{et~al.}, ``Fusion4d: Real-time performance capture of challenging scenes,'' \emph{ACM Transactions on Graphics (ToG)}, vol.~35, no.~4, pp. 1--13, 2016.

\bibitem{huangSCGSSparseControlledGaussian2024}
Y.-H. Huang, Y.-T. Sun, Z.~Yang, X.~Lyu, Y.-P. Cao, and X.~Qi, ``Sc-gs: Sparse-controlled gaussian splatting for editable dynamic scenes,'' in \emph{Proceedings of the IEEE/CVF Conference on Computer Vision and Pattern Recognition}, 2024, pp. 4220--4230.

\bibitem{luitenDynamic3DGaussians2023}
J.~Luiten, G.~Kopanas, B.~Leibe, and D.~Ramanan, ``Dynamic 3d gaussians: Tracking by persistent dynamic view synthesis,'' in \emph{2024 International Conference on 3D Vision (3DV)}.\hskip 1em plus 0.5em minus 0.4em\relax IEEE, 2024, pp. 800--809.

\bibitem{baradObjectcentricReconstructionTracking2024}
\BIBentryALTinterwordspacing
K.~R. Barad, A.~Richard, J.~Dentler, M.~Olivares-Mendez, and C.~Martinez. Object-centric {{Reconstruction}} and {{Tracking}} of {{Dynamic Unknown Objects}} using {{3D Gaussian Splatting}}. [Online]. Available: \url{http://arxiv.org/abs/2405.20104}
\BIBentrySTDinterwordspacing

\bibitem{caiGSPoseCascadedFramework2024}
D.~Cai, J.~Heikkil{\"a}, and E.~Rahtu, ``Gs-pose: Cascaded framework for generalizable segmentation-based 6d object pose estimation,'' \emph{arXiv preprint arXiv:2403.10683}, 2024.

\bibitem{fischerDynamic3DGaussian2024a}
T.~Fischer, J.~Kulhanek, S.~R. Bul{\`o}, L.~Porzi, M.~Pollefeys, and P.~Kontschieder, ``Dynamic 3d gaussian fields for urban areas,'' \emph{arXiv preprint arXiv:2406.03175}, 2024.

\bibitem{wangShapeMotion4D2024}
Q.~Wang, V.~Ye, H.~Gao, J.~Austin, Z.~Li, and A.~Kanazawa, ``Shape of motion: 4d reconstruction from a single video,'' \emph{arXiv preprint arXiv:2407.13764}, 2024.

\bibitem{bansal20204d}
A.~Bansal, M.~Vo, Y.~Sheikh, D.~Ramanan, and S.~Narasimhan, ``4d visualization of dynamic events from unconstrained multi-view videos,'' in \emph{Proceedings of the IEEE/CVF Conference on Computer Vision and Pattern Recognition}, 2020, pp. 5366--5375.

\bibitem{broxton2020immersive}
M.~Broxton, J.~Flynn, R.~Overbeck, D.~Erickson, P.~Hedman, M.~Duvall, J.~Dourgarian, J.~Busch, M.~Whalen, and P.~Debevec, ``Immersive light field video with a layered mesh representation,'' \emph{ACM Transactions on Graphics (TOG)}, vol.~39, no.~4, pp. 86--1, 2020.

\bibitem{cao2023hexplane}
A.~Cao and J.~Johnson, ``Hexplane: A fast representation for dynamic scenes,'' in \emph{Proceedings of the IEEE/CVF Conference on Computer Vision and Pattern Recognition}, 2023, pp. 130--141.

\bibitem{kumar2017monocular}
S.~Kumar, Y.~Dai, and H.~Li, ``Monocular dense 3d reconstruction of a complex dynamic scene from two perspective frames,'' in \emph{Proceedings of the IEEE international conference on computer vision}, 2017, pp. 4649--4657.

\bibitem{ACT}
T.~Z. Zhao, V.~Kumar, S.~Levine, and C.~Finn, ``Learning fine-grained bimanual manipulation with low-cost hardware,'' \emph{arXiv preprint arXiv:2304.13705}, 2023.

\bibitem{wang2023mimicplay}
C.~Wang, L.~Fan, J.~Sun, R.~Zhang, L.~Fei-Fei, D.~Xu, Y.~Zhu, and A.~Anandkumar, ``Mimicplay: Long-horizon imitation learning by watching human play,'' \emph{arXiv preprint arXiv:2302.12422}, 2023.

\bibitem{qin2024langsplat}
M.~Qin, W.~Li, J.~Zhou, H.~Wang, and H.~Pfister, ``Langsplat: 3d language gaussian splatting,'' in \emph{Proceedings of the IEEE/CVF Conference on Computer Vision and Pattern Recognition}, 2024, pp. 20\,051--20\,060.

\bibitem{zhou2024feature3dgs}
S.~Zhou, H.~Chang, S.~Jiang, Z.~Fan, Z.~Zhu, D.~Xu, P.~Chari, S.~You, Z.~Wang, and A.~Kadambi, ``Feature 3dgs: Supercharging 3d gaussian splatting to enable distilled feature fields,'' in \emph{Proceedings of the IEEE/CVF Conference on Computer Vision and Pattern Recognition}, 2024, pp. 21\,676--21\,685.

\bibitem{kerr2023lerf}
J.~Kerr, C.~M. Kim, K.~Goldberg, A.~Kanazawa, and M.~Tancik, ``Lerf: Language embedded radiance fields,'' in \emph{Proceedings of the IEEE/CVF International Conference on Computer Vision}, 2023, pp. 19\,729--19\,739.

\bibitem{Cheng2024YOLOWorld}
T.~Cheng, L.~Song, Y.~Ge, W.~Liu, X.~Wang, and Y.~Shan, ``Yolo-world: Real-time open-vocabulary object detection,'' in \emph{Proc. IEEE Conf. Computer Vision and Pattern Recognition (CVPR)}, 2024.

\bibitem{kratimenos2023dynmf}
A.~Kratimenos, J.~Lei, and K.~Daniilidis, ``Dynmf: Neural motion factorization for real-time dynamic view synthesis with 3d gaussian splatting,'' \emph{arXiv preprint arXiv:2312.00112}, 2023.

\bibitem{huang2023earl}
B.~Huang, J.~Yu, and S.~Jain, ``Earl: Eye-on-hand reinforcement learner for dynamic grasping with active pose estimation,'' in \emph{2023 IEEE/RSJ International Conference on Intelligent Robots and Systems (IROS)}.\hskip 1em plus 0.5em minus 0.4em\relax IEEE, 2023, pp. 2963--2970.

\bibitem{wen2024foundationpose}
B.~Wen, W.~Yang, J.~Kautz, and S.~Birchfield, ``Foundationpose: Unified 6d pose estimation and tracking of novel objects,'' in \emph{Proceedings of the IEEE/CVF Conference on Computer Vision and Pattern Recognition}, 2024, pp. 17\,868--17\,879.

\bibitem{weyrich2007ray-splat}
T.~Weyrich, S.~Heinzle, T.~Aila, D.~B. Fasnacht, S.~Oetiker, M.~Botsch, C.~Flaig, S.~Mall, K.~Rohrer, N.~Felber, \emph{et~al.}, ``A hardware architecture for surface splatting,'' \emph{ACM Transactions on Graphics (TOG)}, vol.~26, no.~3, pp. 90--es, 2007.

\bibitem{zwicker2001ewa}
M.~Zwicker, H.~Pfister, J.~Van~Baar, and M.~Gross, ``Ewa volume splatting,'' in \emph{Proceedings Visualization, 2001. VIS'01.}\hskip 1em plus 0.5em minus 0.4em\relax IEEE, 2001, pp. 29--538.

\bibitem{CLIP}
A.~Radford, J.~W. Kim, C.~Hallacy, A.~Ramesh, G.~Goh, S.~Agarwal, G.~Sastry, A.~Askell, P.~Mishkin, J.~Clark, \emph{et~al.}, ``Learning transferable visual models from natural language supervision,'' in \emph{International conference on machine learning}.\hskip 1em plus 0.5em minus 0.4em\relax PMLR, 2021, pp. 8748--8763.

\bibitem{SAM}
A.~Kirillov, E.~Mintun, N.~Ravi, H.~Mao, C.~Rolland, L.~Gustafson, T.~Xiao, S.~Whitehead, A.~C. Berg, W.-Y. Lo, \emph{et~al.}, ``Segment anything,'' in \emph{Proceedings of the IEEE/CVF International Conference on Computer Vision}, 2023, pp. 4015--4026.

\bibitem{fang2020graspnet}
H.-S. Fang, C.~Wang, M.~Gou, and C.~Lu, ``Graspnet-1billion: A large-scale benchmark for general object grasping,'' in \emph{Proceedings of the IEEE/CVF conference on computer vision and pattern recognition}, 2020, pp. 11\,444--11\,453.

\bibitem{foundationposewen2024}
J.~K. S.~B. Bowen~Wen, Wei~Yang, ``{FoundationPose}: Unified 6d pose estimation and tracking of novel objects,'' in \emph{CVPR}, 2024.

\bibitem{handeye_calib}
K.~Daniilidis, ``Hand-eye calibration using dual quaternions,'' \emph{The International Journal of Robotics Research}, vol.~18, no.~3, pp. 286--298, 1999.

\bibitem{sfm}
J.~L. Schonberger and J.-M. Frahm, ``Structure-from-motion revisited,'' in \emph{Proceedings of the IEEE conference on computer vision and pattern recognition}, 2016, pp. 4104--4113.

\bibitem{shorinwa2024splat}
O.~Shorinwa, J.~Tucker, A.~Smith, A.~Swann, T.~Chen, R.~Firoozi, M.~D. Kennedy, and M.~Schwager, ``Splat-mover: Multi-stage, open-vocabulary robotic manipulation via editable gaussian splatting,'' in \emph{8th Annual Conference on Robot Learning}, 2024.

\bibitem{liu2024aligning}
Y.~Liu, W.~Chen, Y.~Bai, J.~Luo, X.~Song, K.~Jiang, Z.~Li, G.~Zhao, J.~Lin, G.~Li, \emph{et~al.}, ``Aligning cyber space with physical world: A comprehensive survey on embodied ai,'' \emph{arXiv preprint arXiv:2407.06886}, 2024.

\end{thebibliography}
